\documentclass[conference]{IEEEtran}
\pdfoutput=1

\usepackage{amsmath,amsfonts,amsthm} % Math packages
\usepackage{graphicx}
\usepackage{subcaption}
\usepackage{float}
\usepackage{pbox}
\usepackage{hyperref}
\usepackage{listings}
\usepackage{pifont}
\usepackage{xcolor}
\usepackage{courier}
\usepackage{soul}

\begin{document}
\title{Genetic Algorithms for multimodal optimization:\\ A review}

\author{\IEEEauthorblockN{Noe Casas}
\IEEEauthorblockA{ Email: research@noecasas.com}}

\maketitle

\begin{abstract}
%\boldmath
In this article we provide a comprehensive review of the different evolutionary algorithm
techniques used to address multimodal optimization problems, classifying them
according to the nature of their approach. On the one hand there are algorithms that
address the issue of the early convergence to a local optimum by differentiating
the individuals of the population into groups and limiting their interaction, hence
having each group evolve with a high degree of independence. On the other hand other
approaches are based on directly addressing the lack of genetic diversity of the
population by introducing elements into the evolutionary dynamics that promote
new niches of the genotypical space to be explored. Finally, we study multi-objective
optimization genetic algorithms, that handle the situations where multiple criteria
have to be satisfied with no penalty for any of them. Very rich literature has
arised over the years on these topics, and we aim at offering an overview of the
most important techniques of each branch of the field.
\end{abstract}


% \fi
\IEEEpeerreviewmaketitle

\section{Introduction}

Genetic Algorithms aim at exploring the genotypical space to find an individual whose associated
phenotype optimizes a prefefined fitness function. When the fitness function presents multiple
local optima, the problem is said to be multimodal optimization.\\

The desired behaviour of a GA when applied to such a type of problem is not to get stuck
in local optima, but find the global optimum of the function. This very same concept was
introduced in the context of biological evolution by Sewall Wright in \textit{Fitness Landscapes}
in 1932 in \cite{wright1932roles}:

\begin{quote}
In a rugged field of this character, selection will easily carry the species to the
nearest peak, but there may be innumerable other peaks which are higher but
which are separated by ‘valleys’. The problem of evolution as I see it
it is that of a mechanism by which the species may continually find
its way from lower to higher peaks in such a field. In order for this to occur,
there must be some trial and error mechanism on a grand scale by which the
species may explore the region surrounding the small portion of the field
which it occupies
\end{quote}

This type of scenarios require the GA to develop strategies to cover all the genotypic
space without converging to a local optimum. Throughout this article we explore the different
techniques applied to genetic algorithms to improve their effectiveness in multimodal fitness problems.\\

Multimodal optimization problems should not be confused with multiobjective optimization:
while multimodal optimization tries to find the optimum of a single
fitness function that has multiple local optima, multi-objective optimization tries to find the balance at optimizing
several fitness functions at the same time. Nevertheless, they have many commonalities, and
concepts from each world can be applied to the other. For instance, current multi-objective
genetic algorithms tend to make use of multimodal optimization mechanisms in one of their steps,
normally \textit{crowding} or \textit{fitness sharing}(see section \ref{sec:fitnesssharing}), and
also, recent reinterpretations of multi-objective genetic algorithms allow to explicitly
hadle multimodal single-objective optimization problems.\\

In section \ref{sec:stateoftheart} we explore the different groups of techniques, covering first the
approaches that structures the population so the the individuals are assigned to subpopulations with limited cross-interation
(section \ref{sec:structuredpopulation}) and then the algorithms that condition the evolutionary dynamics
to promote diversity, aiming at avoiding early convergence to a local optimum. We also explore
multi-objective optimization algorithms (\ref{sec:multiobjective}), focusing on their aspects related
to multi-modality. After the glance at the available techniques, section \ref{sec:discussion} provide a brief discussion
their pros and cons. Finally, our conclusions are presented in section \ref{sec:conclusions}

\section{State of the art} \label{sec:stateoftheart}

Throughout this section we explore the different genetic algorithm branches devoted to multimodal problem
optimization, describing the most remarkable representatives of each of the approaches.\\

It should be noted that in the literature, many of the algorithms described in the following
sections are labeled under the term \textit{niching methods}. The uses of such a wording
are diverse, normally comprising GAs based on the spatial distribution of the population
(section \ref{sec:spatialdistribution}) and some GAs based on the explicit control of the
diversity within the population (\ref{sec:diversityenforcing}). However, due to the broad use of
the word, we shall not use it, in order to avoid misunderstandings.

\subsection{Structured Population GAs (implicit diversity promotion)} \label{sec:structuredpopulation}

Structured Population Genetic Algorithms do not explicitly measure and enforce diversity, but impose
certain constraints to the population aiming at regulate its dynamics. The two subgroups among this
type of algorithms are those explicitly partitioning the population and those inducing measures
that lead individuals to cluster into subpopulations. Both families are studied in the following subsections.

\subsubsection{Algorithms based on the Spatial Segregation of One Population} \label{sec:spatialsegregation}

Evolutionary Algotihms relying on spacial segregation usually divide the population into
completely segragated groups that at certain points in time retrofit their genetic material.
This aims at avoiding the homogenicity of panmitic approaches by keeping several homogenous
groups and mixing them at controlled intervals, hence profitting from the good local optima
found by each subgroup.\\

\begin{itemize}

\item \textbf{Island Model GAs}  (\textit{aka} Parallel GAs, Coarse-grained GAs) manage
subpopulations (islands, demes), each one evoling separately with its
own dynamics (e.g. mutation rate, population size), but at certain points they exchange some
individuals (i.e. the new genetic operator: migration). The algorithm can be subject of different
design decissions (\cite{izzo12generalized}): \textbf{number of islands} and \textbf{migration topology} among them
(individuals from which islands can migrate to which islands), \textbf{synchronism} of the migration (asynchronous
scales better and profits from underlying hardware parallelism, but is non-deterministic, non-repeatable
and difficult to debug), migration frequency, whether the migration is initiated by the source or by
the destination and the \textbf{migration selection} and \textbf{migration replacement} policies.

%\hl{TODO: mimick Figure1 of izzo12}

\item \textbf{Spatially-Dispersed GAs} (\cite{dick03spatiallydispersed}) associate a two dimensional
coordinate to every individual (initial positions are assigned at random), having offspring placed randomly but close to
the first of the parents. Mating is only allowed with individuals that are located within certain
\textit{visibility radius}. These dynamics lead to the progressive spread and accumulation of the originally
randomly distributed individuals into clusters (demes) that resemble subpopulations. The value of the aforementioned
visibility radius is not important, as the populations spread according to its scale  
without performance penalties.

%\hl{TODO: mimick Figure1 of dick03}

\end{itemize}

\subsubsection{Algorithms based on Spatial Distribution of One Population} \label{sec:spatialdistribution}

Genetic Algorithms belonging to this family do not impose hard divisions among groups of individuals, but
induce their clustering by means of constraints in their evolutive dynamics. Their most remarkable approaches are:

\begin{itemize}

\item \textbf{Diffusion Model} (\cite{husband94distributed, white97diffusion}) keeps two subpopulations,
said to be of different species. The individuals from both populations are spread over the same two-dimensional
toroidal grid, ensuring each cell contains only one member from each population. Mating is restricted
to individuals from the same species within the neighbouring cells with a fitness-proporctionate scheme.
Replacement follows the opposite approach from mating, that is, offspring probabilistically replaces their parents
in the neighbourhood. Both populations compete to be the fittest, hence they co-evolve but do not mix
with the other species.

%\hl{TODO: mimick figure 4 from husband94distributed, but without perspective}

\item \textbf{Cellular GAs} (cGA) (\cite{alba08cellular}) is the name of the family of GAs evolved from the
Diffusion Model. They also adjust the selection pressure and have the concept of neighbourhoods, but
improve the base idea on several different directions. Some of the remarkable contributions are:

\begin{itemize}

\item \textbf{Terrain-Based Genetic Algorithms} (TBGA) (\cite{gordon99terrain,gordon04terrain}) is a self-tuning version
of cGA, where each grid cell of the two-dimensional world is assigned a different combination of parameters.
They then evolve separately, each cell mating with their up, down, left and right neighbours. This algorithm
can be used not only to address the optimization problem itself, but to find a set of suitable parameter
values to be used in a normal cGA. In fact, the authors admit that a normal cGA using the parameters found
by their TBGA performs better than the TBGA itself.

\item \textbf{Genetic and Artificial Life Environment} (GALE) (\cite{llora01knowledge}) offers the concept
of empty cells, where neighbouring offspring are placed. If no empty cells are present after breeding a cell,
new individuals replace worst performing individuals from their original neighbourhood. This algorithm also
presents fitness sharing (see section \ref{sec:fitnesssharing}).

\item \textbf{Co-evolutionary} approaches like in \cite{hillis91coevolving}, an improvement over \textit{sorting networks},
where two species (referred to as hosts-parasites or prey/predators). Hosts are meant to sort some input data, while
parasites represent test data to be supplied to a host as input. The fitness of each group is opposed to the
other group: the fitness of the hosts depends on how many test cases (i.e. parasites) an individual has succeeded in
sorting, while the fitness of the parasites depends on how many times it made a host fail sorting.

\item \textbf{Multi-objective} variations of cGA, namely cMOGA and MOCell, which are addressed in section \ref{sec:multiobjective}.

\end{itemize}

\end{itemize}

\subsubsection{Algorithms imposing other mating restrictions}

These algorithms impose mating restrictions based on other criteria, normally mimicking
the high level dynamics of existing real-world environments. The most remarkable
ones present in the literature are:

\begin{itemize}
\item Multinational Evolutionary Algorithms (\cite{ursem99multinational}): divide the
world into nations and partition the population among them, also having different
\textit{roles} within each nation, namely politicians, police and normal people. Their
interaction and mating dynamics are defined by pre-established social rules.
\item Religion-Based Evolutionary Algorithms (\cite{thomsen00religion}): assigns each individual to a different
religion and defines genetic operators for converting between religions. Mating is hence
restricted to individuals with the same beliefs.
\item Age Structure GAs (\cite{kubota94genetic}) define the lifecycle of individuals
and constrain the mating to individuals in the same age group.
\end{itemize}

\subsection{Diversity Enforcing Techniques} \label{sec:diversityenforcing}

The main trait of this group of algorithms is that they define a measure of the population diversity
distribution over the genotypical space and act upon local accumulaions of individuals, favouring
heir migration to new niches.

\subsubsection{Fitness sharing} \label{sec:fitnesssharing}

Fitness Sharing GAs (\cite{goldberg87sharing}) are based on having individual's
\textit{fitness points} shared with their
neighbours. The \textit{neighbourhood} is defined as the individuals within certain radius $\sigma_{share}$
over a established distance metric (e.g. euclidean distance, Hamming distance). This way,
the new fitness $F'$ of an individual $i$ is calculated based on its distance $d$ to every
neighbour $j$ as:

\begin{equation}
F'(i) = \dfrac{F(i)}{\sum_j sharingfunction(d(i,j))}
\end{equation}

where $sharingfunction$ receives as input the distance between two individuals and is
computed as:

\begin{equation}
sharingfunction(d) = \left\{
  \begin{array}{lr}
    1 - (d/\sigma_{share})^\alpha : if d \leq \sigma_{share} \\
    0 : otherwise
  \end{array}
\right.
\end{equation}

Having $\alpha$ define the shape of the sharing function (i.e. $\alpha=1$ for lineal sharing).\\

This way, the convergente to a single area of the fitness landscape is discouraged 
by pretending there are limited resources there. The more individuals try to move in, the more
neighbours the fitness have to be shared with. Hence, for individuals in crowded areas, eventually another
region of the fitness space becomes more attractive. Ideally, the algorithm stabilizes 
at a point where an appropriate representation of each niche is maintained.

\subsubsection{Clearing}

Clearing GAs (\cite{petrowski96clearing}) divide the population in
subpopulations according to a dissimilarity measure (e.g. Hamming distance). For each
subpopulation, in the selection phase the fittest individual is considered
\textit{the winner} (normally referred to as \textit{the dominant individual}). Then
the other members of the subpopulation have their dissimilarity to the winner calculated.
If such the distance of an individual of the subpopulation to its winner is greater than certain
threshold (the \textit{clearing radius}), it gets its fitness set to zero (i.e. \textit{it gets cleared}).
After the whole population has been processed, the subpopulations are recalculated again
based on the very same clearing radius.

\subsubsection{Crowding} \label{sec:crowding}

Crowding GAs associate to every individual breeded in
the current generation with another individual from the parent generation (pairing phase) and only
keep one of the two in the population (replacement phase). The association is established
based on genotypical similarity criteria (e.g. Manhattan distance, Euclidean distance).
This approach favors the growth of individuals around underpopulated regions of the solution space and
penalizes overcrowded areas because only the similar individuals get replaced.\\

In the original algorithm formulation, De Jong used a Crowding Factor parameter (CF) (section
4.7 of \cite{dejong75analysis}) to specify the size of the sample of individuals initially selected
at random as candidates to be replaced by a particular offspring, among which only one shall finally be chosen
based on fitness. He found problems in the original formulation of the algorithm, as it failed to
prevent genetic drift in many cases.\\

Mahfoud improved on the algorithm by identifying several weak
points, most remarkably focusing on maintaining global diversity
(\cite{mahfoud92crowding,mahfoud93simpleanalytical,mahfoud95niching}) and addressing them by introducing a different diversity measure
that favoured niching, namely the number of \textit{peaks} maintained by the population. With
the new measures, Mahfoud re-evaluated De Jong's mislead conclusions (i.e. that CF higher than 1
led to genetic drift) and reformulated the algorithm to only use the individual's parents as
candidates for replacement (hence reducing drastically the computational complexity). This variation
is called \textbf{Deterministic Crowding}.\\

Mengshoel (\cite{mengshoel99probabilistic}) proposed a variation called \textbf{Probabilistic Crowding}
in which the selection criteria for individuals to be replaced is not fitness-proportionate
but random, hence favouring the conservation of low-fitness individuals and avoiding genetic
drift toward the high-fitness niche.

\subsection{Multi-objective evolutionary algorithms} \label{sec:multiobjective}

Multi-Objective Optimization problems are characterized by the need to find proper \textit{trade-offs}
among different criteria, each of them quantified by means of an objective function, formally (\cite{coello06evolutionary}):

\begin{quotation}
A general \textbf{Multi-objective Optimization Problem} (MOP) is defined as minimizing (or maximizing)
$F(x) = (f_1 (x), ... , f_k (x))$ subject to $g_i (x) \leq 0$, $i = {1, ... ,m}$, and
$h_j(x) = 0$, $j = {1, ... , p}$. An MOP
solution minimizes (or maximizes) the components of a vector $F(x)$ where $x$ is
a n-dimensional decision variable vector $x = (x_1 , ... , x_n )$ from some universe
$\Omega$. It is noted that $g_i (x) \leq 0$ and $h_j (x) = 0$ represent constraints that must be
fulfilled while minimizing (or maximizing) $F(x)$ and $\Omega$ contains all possible $x$
that can be used to satisfy an evaluation of $F(x)$.
\end{quotation}

The evaluation function, $F : \Omega \rightarrow \Lambda$, maps from the
decision variable space $\bar{x} = (x_1 , ... , x_n )$ to the objective function
space $y = f(\bar{x}), ... , f_k(\bar{x}))$ \footnote{In this equation we use the $\bar{x}$ notation
to clarify the vectorial nature of the parameter of $f_i$, but we will not use it in the
rest of the report}.\\

The cathegorization
\textit{Multi-objective optimization} normally refers to approaches that are defined in terms of \textit{Pareto
Optimality} (\cite{coello06evolutionary}):

\begin{quotation}
A solution $x \in \Omega$ is said to be \textbf{Pareto Optimal} with respect to $\Omega$ if and only if
there is no $x \in \Omega$ for which $v=F(x)=(f_1 (x), ... , f_k (x))$ dominates
$u = F (x) = (f_1 (x), ... , f_k (x))$.
\end{quotation}

Where a vector $u$ s said to \textit{dominate} another vector $v$ if there is a subset of $f_i(x)$ for
which $u$ is (assuming minimization) partially less than $v$, that is $\exists i : u_i < v_i$.\\

This means that $x^*$ is Pareto optimal if there exists no vector which would decrease some criterion
without causing a simultaneous increase in at least one other criterion (again assuming minimization).\\

When we plot all the objective function vectors that are nondominated, the obtained curve is usually
referred to as the \textbf{Pareto front}.\\ 

A \textbf{Multi-objective Optimization Evolutionary Algorithm} (MOEA) consists of the application of GAs to
a MOE. The mechanism of a MOEA is the same as a normal GA. Their only difference is that, instead of a single fitness function, MOEAs
compute $k$ fitness function and then perform on them a transformation in order to obtain a single
measure, which is then used as the fitness in normal GAs.  
At each generation MOEAs output is the current set of Pareto optimal solution (i.e. the Pareto front)
\footnote{Some MOEAs make use of a secondary population acting as an \textit{archive}, where they store all
the nondominated solutions found through the generations}.\\

There exist multiple variations of MOEAs, each of them differring in either the way they combine the
individual objective functions into the fitness value (\textit{a priori techniques})or the post-processing they do to
ensure Pareto optimality.\\

Deb recently proposed a MOEA (\cite{deb10finding,saha10bicriterion}) for addressing single-objective
multimodal optimization problems. This algorithm defined a suitable second objective and added it to the
originally single objective multimodal optimization problem, so that the multiple solutions form a
pareto-optimal front. This way, the single-objective multimodal optimization problem \textit{turned artificially into a MOP}
can be solved for its multiple solutions using the MOEA.

\section{Discussion} \label{sec:discussion}

One of the most attractive traits of spatial segregation GAs is that each of the population
subgroups can be evolved in parallel, hence making them suitable to profit from parallel
architectures such as multicore or supercomputing facilities.\\

Another potentially attractive chracteristic of this type of algorithms is that they are
to some degree independent on the optimization algorithm. This enables to use different
optimization algorithms (not constraining to GAs) to each island (\cite{izzo12generalized}).

However, a significant concern about them is that they need to be carefully tuned in
order to perform well. For instance, \textit{Fitness sharing} (section \ref{sec:fitnesssharing}) needs
to manually set the niche radius, and the algorithm is quite sensitive to this choice. Failure to properly
tune the algorithm parameters normally implies
performing significantly under that of an equivalent panmitic implementation (\cite{dick03spatiallydispersed}).
In this regard, Spatially-Dispersed GAs require less configuration tuning than Island Model GAs.\\

Some self-adapting options are interesting in that they do not need such configuration tuning
at all. However, many times these algorithms perform worse than their equivalent fine tuned non-adaptive
version (\cite{gordon99terrain,gordon04terrain}).

One of the most significant problems among the reviewed techniques is the one suffered by those that rely on
structuring the population (section \ref{sec:structuredpopulation}), which cannot
assure an improvement on the solution space covered because they are not based on a measure of
the distribution of the diversity (\cite{artyushenko09analysis}). This way, these algorithms offer good performance for some
problems, but worse performance than panmitic approaches, with no apparent reason. Moreover,
given their loose relation to diversity control, it is often impossible to diagnose or fix
the root cause of the algorithm underperforming for certain problem.\\

Most of the sources in the literature of the last years tend to agree that \textit{Crowding} (section \ref{sec:crowding})
is effective for any multimodal optimization problem. Nevertheless, promising MOEAs (\cite{deb10finding,saha10bicriterion}) 
may also play an important role in the upcoming years.

\section{Conclusions} \label{sec:conclusions}

In most real-world optimization problems, the fitness landscape is unknown to us. This means that
there is a non-negligible chance that it is multimodal. Failing to acknowledge so -and act accordingly-
may likely result in the optimization to converge too early to a local optimum.\\

From the reviewed techniques, the only one with quorum among the scientific community regarding
general effectiveness is Crowding. All other options have proven valuable for many concrete
problems, but they certainly exhibit suboptimal performance compared to panmitic approaches
for some other problems.\\

This tells us that selecting the appropriate multimodal optimization genetic algorithm cannot be
addressed \textit{a priori}, but has to undergo a trial-error process, driven by the intuition
of the researchers to choose an approach that has proven effective for seemingly analogous or similar
problems.



\bibliographystyle{IEEEtran}
\bibliography{biblio}

\end{document}